\documentclass[letterpaper]{article} 
\usepackage{aaai2026}  
\usepackage{times}  
\usepackage{helvet}  
\usepackage{courier}  
\usepackage[hyphens]{url}  
\usepackage{graphicx} 
\usepackage{natbib}  
\usepackage{caption} 
\usepackage{algorithm}
\usepackage{algorithmic}

\usepackage{newfloat}
\usepackage{listings}
\DeclareCaptionStyle{ruled}{labelfont=normalfont,labelsep=colon,strut=off} 
\floatstyle{ruled}
\newfloat{listing}{tb}{lst}{}
\floatname{listing}{Listing}
\title{CoEvo: Continual Evolution of Symbolic Solutions Using Large Language Models}
\author{
    Ping Guo\textsuperscript{\rm 1}, Qingfu Zhang\textsuperscript{\rm 1}\thanks{Corresponding author: qingfu.zhang@cityu.edu.hk}, Xi Lin\textsuperscript{\rm 1}
}
\affiliations{
    \textsuperscript{\rm 1}City University of Hong Kong\\

    No. 83 Tat Chee Road, Hong Kong SAR, China\\
}

\usepackage{bibentry}

\usepackage{subfigure}
\usepackage{multirow}
\usepackage{booktabs}
\usepackage{color}
\usepackage[table]{xcolor}
\graphicspath{{figs/}} 
\definecolor{light-gray}{gray}{0.9}

\usepackage{xspace}
\newcommand{\bheading}[1]{{\vspace{0.25\baselineskip}\noindent{\textbf{#1}}}}

\newcommand{\figref}[1]{\figurename~\ref{#1}}

\newcommand{\tabref}[1]{\tablename~\ref{#1}}

\begin{document}

\maketitle

\begin{abstract}
  The discovery of symbolic solutions—mathematical expressions, logical rules, and algorithmic structures—is fundamental to advancing scientific and engineering progress.
  However, traditional methods often struggle with search efficiency and fail to integrate knowledge effectively.
  While recent large language model-based (LLM-based) approaches have demonstrated improvements in search efficiency, they lack the ability to continually refine and expand upon discovered solutions and their underlying knowledge, limiting their potential for \textit{open-ended innovation}.
  To address these limitations, we introduce CoEvo, a novel framework that leverages large language models within an evolutionary search methodology to continually generate and refine symbolic solutions. CoEvo integrates a dynamic knowledge library, enabling open-ended innovation of solutions through effective knowledge management. Additionally, CoEvo leverages multiple representations of solutions—including natural language, mathematical expressions, and code—to further enhance search efficiency.
  By combining the reasoning capabilities of LLMs with the exploratory power of evolutionary algorithms, CoEvo significantly improves the efficiency and scope of symbolic discovery.
  Our experimental results demonstrate that this method not only enhances the efficiency of searching for symbolic solutions but also supports the ongoing discovery process, akin to human scientific endeavors. This study represents a first effort in conceptualizing the search for symbolic solutions as a lifelong, iterative
  process, marking a significant step towards harnessing LLMs in the perpetual pursuit of scientific and engineering breakthroughs.
  Our code is available at \url{https://github.com/pgg3/CoEvo}.
\end{abstract}

\section{Introduction}
The pursuit of symbolic solutions—mathematical models, logical rules, and algorithmic structures—lies at the heart of scientific and engineering progress, underpinning both theoretical frameworks and practical applications~\cite{wigner:1990:unreasonable,newell:1980:physical}.
From designing complex engineering systems to formulating new scientific theories, the discovery of such solutions drives innovation and technological advancement~\cite{gielen:2012:symbolic,schmidt:2009:distilling}.
For instance, symbolic solutions like Intellectual Property (IP) blocks are essential for optimizing system performance and accelerating development cycles~\cite{brown:2000:fundamentals,wolf:2002:modern}.
However, the process of discovering these solutions is often hindered by the difficulties of navigating vast representation spaces and integrating new knowledge into the search process.

A promising approach to advancing symbolic discovery involves emulating the open-ended, iterative nature of human scientific and engineering endeavors, where solutions and foundational knowledge co-evolve over time.
This approach has the potential to unlock innovative ideas that are inaccessible through conventional methods~\cite{stanley:2017:open,lenat:1983:eurisko}.
In artificial intelligence, open-ended exploration has led to significant breakthroughs by establishing environments where algorithms endlessly generate and refine solutions without the constraints of predefined goals~\cite{lehman:2011:abandoning,faldor:2024:omniepic}.
This paradigm mirrors the iterative essence of human creativity and scientific discovery, where each insight sparks further questions and exploration~\cite{boden:2004:creative}.

Recent advances in large language models (LLMs) have demonstrated their ability to automate problem-solving tasks across diverse domains, including code generation and scientific reasoning~\cite{openai:2023:gpt4,liu:2024:systematic,ifargan:2024:autonomous}.
However, their application to open-ended symbolic discovery—where solutions and knowledge continually evolve—remains underexplored.
While LLMs incorporate baseline human knowledge, they struggle to integrate newly discovered insights and generate novel solutions in an open-ended manner,
not to mention traditional methods such as evolutionary algorithms~\cite{cranmer:2023:pysr} and deep learning approaches~\cite{biggio:2021:neural,kamienny:2022:end}.
Although effective in specific contexts, these methods often fall short for achieving the open-ended innovation characteristics of human scientific endeavors.

Efforts to enhance LLMs for symbolic discovery have focused on improving their problem-solving capabilities through techniques such as retrieval-augmented generation (RAG)~\cite{huang:2024:surveyrag} and domain-specific fine-tuning~\cite{roziere:2023:code}.
While these approaches enhance search efficiency, they primarily enable LLMs to reuse prior knowledge rather than create or refine new knowledge.
This raises a critical question:

\emph{Can LLMs uncover new knowledge rather than merely replicate existing information? In addition, can LLMs summarize and evolve knowledge to support an open-ended search for symbolic solutions akin to human endeavors?}

To address these challenges, we introduce CoEvo, a novel framework that combines LLMs with an evolutionary search methodology to continually generate and refine symbolic solutions alongside underlying knowledge.
CoEvo incorporates a dynamic knowledge library and multiple representation spaces, enabling the open-ended evolution of solutions across diverse formats, including natural language, mathematical expressions, and code.
This approach tackles two key challenges: (1) the management of evolving knowledge and (2) the exploration of diverse representation spaces in an open-ended manner.
To our knowledge, this represents the first effort to frame symbolic discovery as a continual, open-ended process powered by LLMs.

Our contributions are as follows:
\begin{itemize}
    \item We make a first attempt to extract knowledge and apply it in the endless search for symbolic solutions for scientific and engineering challenges. We believe that finding symbolic solutions in domains such as scientific discovery should be perceived as a continual open-ended process.
    \item We propose a framework for harvesting and applying knowledge to search for symbolic solutions in multiple search spaces, with the understanding that the knowledge is continually evolving.
    \item Our extensive experimental outcomes demonstrate that incorporating the newly generated knowledge enables a lifetime of continual searching for symbolic solutions in scientific and engineering domains.
\end{itemize}

\section{Background}

\subsection{Symbolic Regression}
Symbolic regression has been widely employed to uncover mathematical equations that capture underlying relationships in datasets~\cite{udrescu:2019:aifeynman}.
Before the rise of LLMs, techniques in this field were broadly categorized into three main approaches: search-based, learning-based, and hybrid methods~\cite{cranmer:2023:pysr}.

Conventional search-based approaches primarily employ evolutionary algorithms to explore the space of equation structures and parameters.
These methods rely on carefully designed solution representations, such as expression trees, to iteratively evolve candidate solutions~\cite{schmidt:2009:distilling,cranmer:2023:pysr}.
However, designing such representations is challenging, and their efficiency in traversing the search space remains uncertain~\cite{kronberger2024inefficiency}.

With the advances in transformer-based architectures, learning-based models have become increasingly prominent in symbolic regression.
This shift has also led to hybrid methods that combine the strengths of both search-based and learning-based approaches~\cite{biggio:2021:neural,kamienny:2022:end}.
While these methods have addressed some efficiency concerns, they still struggle to effectively incorporate prior knowledge, which could further enhance both the efficiency and interpretability of the discovered solutions.

\begin{figure}[t]
    \centering
    \includegraphics[width=0.38\textwidth]{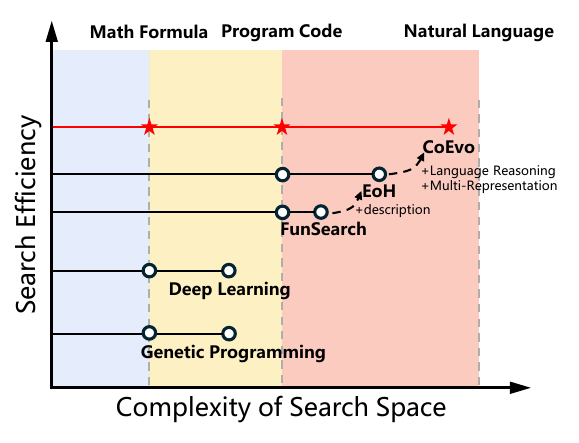}
    \vspace{-1\baselineskip}
    \caption{Conceptual comparison of symbolic discovery methods across search spaces of increasing complexity and knowledge richness. Traditional approaches (pre-LLM) operate in constrained mathematical/code spaces. LLM-based methods (FunSearch, LLM-SR) leverage iterative evolution in code space, while EoH incorporates natural language heuristics. CoEvo (proposed) fully exploits LLMs' reasoning in natural language space for open-ended evolution.}
    \label{fig:comparison}
\end{figure}

\begin{figure*}[t]
    \centering
    \includegraphics[width=0.98\textwidth]{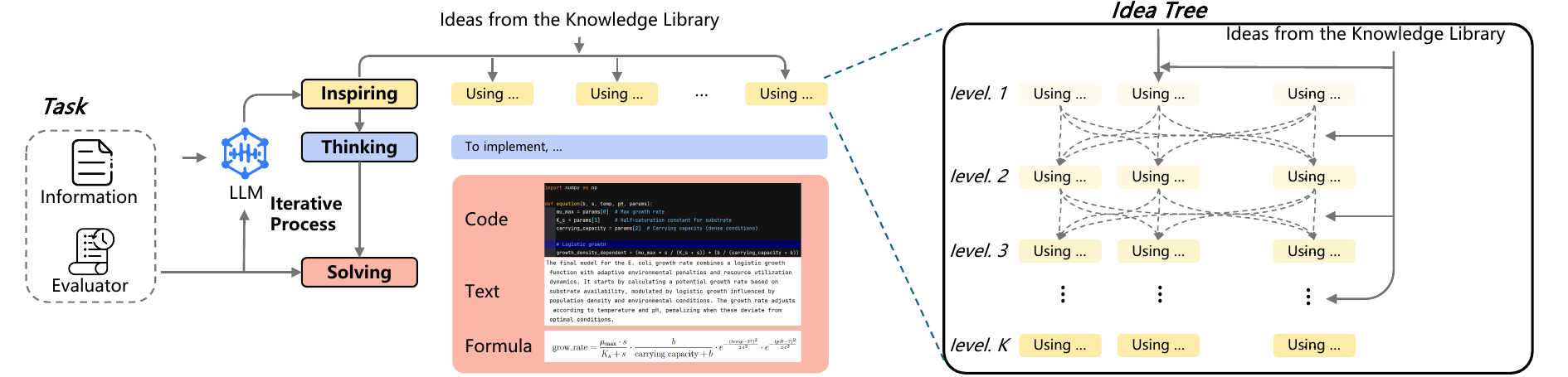}
    \caption{Three-step LLM-driven solution generation (inspiring/thinking/solving) via an idea tree: roots evolve through evaluator-guided refinement over levels, output in multi-format representations.}
    \vspace{-1.25\baselineskip}
    \label{fig:llm_think}
\end{figure*}

\subsection{LLMs and Symbolic Discovery}
LLMs have been increasingly employed in scientific discovery tasks due to their ability to process and generate natural language, code, and mathematical expressions~\cite{openai:2023:gpt4,abramson:2024:accurate}.
Initially, LLMs were applied to discover mathematical equations and algorithms through iterative evolution of solutions implemented in Python code~\cite{liu:2024:systematic,paredes:2024:funsearch,liu:2024:eoh}.
This approach, which equips LLMs with a systematic evaluator and enables interaction via program code, has demonstrated promising results in deriving symbolic solutions. For instance, LLM-SR~\cite{shojaee:2024:llmsr} adopts the methodology of FunSearch~\cite{paredes:2024:funsearch} to iteratively evolve solutions, which are then forwarded to specialized evaluators.

The state-of-the-art method, LLM-SR, has achieved notable success in symbolic regression tasks, validating the effectiveness of iterative evolution and establishing an evaluation framework for such tasks. However, it assumes static knowledge and generates solutions in a single format, failing to fully exploit the reasoning capabilities of LLMs to derive new knowledge.

\figref{fig:comparison} presents a conceptual comparison of existing methods. We categorize the complexity of the search space into three levels: \textit{1)} mathematical formula space, \textit{2)} code space, and \textit{3)} natural language space. As the complexity of the search space increases, so does the richness and complexity of the knowledge it encompasses.

The figure contrasts the search efficiency of conventional symbolic regression methods with LLM-based approaches. Prior to the advent of LLMs, both search-based and learning-based methods were confined to the mathematical formula and code spaces, with limited capacity to leverage broader knowledge. With the emergence of LLMs, the search space has expanded to include natural language. Applications such as FunSearch~\cite{paredes:2024:funsearch} and LLM-SR~\cite{shojaee:2024:llmsr} have demonstrated the efficacy of iterative evolution in symbolic regression tasks. Another approach, EoH~\cite{liu:2024:eoh}, also employs iterative evolution but further capitalizes on the linguistic capabilities of LLMs by incorporating natural language heuristics. Our proposed method, CoEvo, advances this further by harnessing the reasoning abilities of LLMs to derive new knowledge and evolve solutions in an open-ended manner.

\section{CoEvo: Continual Evolution of Symbolic Solutions using LLMs}

\begin{figure}[t]
    \centering
    \includegraphics[width=0.45\textwidth]{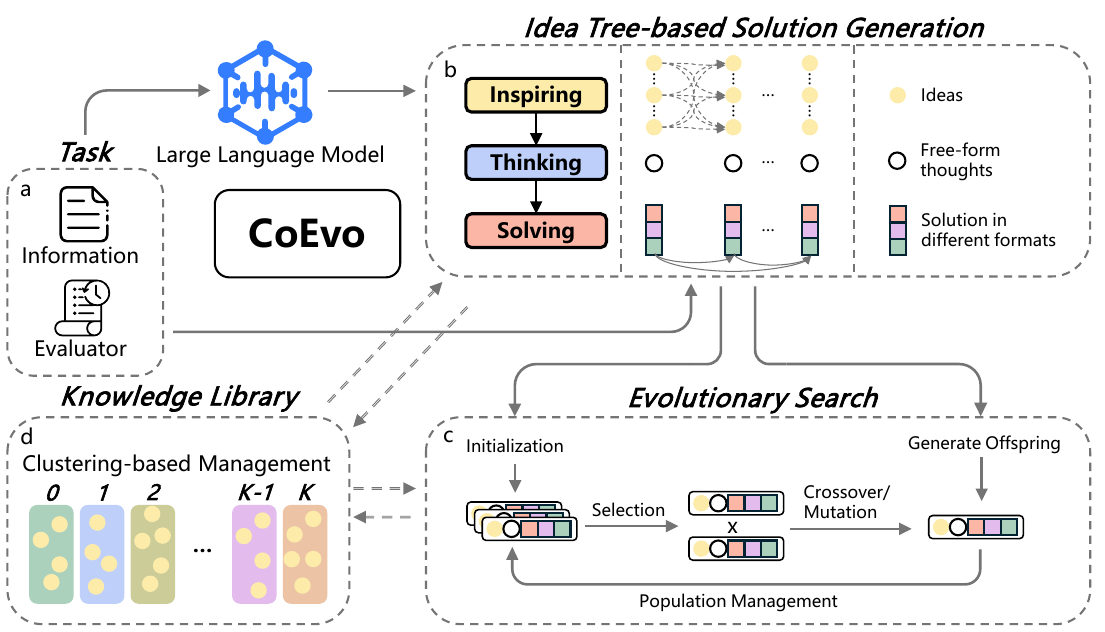}
    \vspace{-0.75\baselineskip}
    \caption{\textbf{An overview of CoEvo.} \textbf{(a)} Task of interest. \textbf{(b)} Tree-based solution generation for generating of a single solution in different formats. \textbf{(c)} Evolutionary search of solutions. \textbf{(d)} Knowledge library for storing and retrieving knowledge pieces.}
    \label{fig:coevo}
\end{figure}

\subsection{Overview}
CoEvo is a framework designed to facilitate a continuous and open-ended search process for symbolic solutions while effectively managing underlying knowledge, as illustrated in \figref{fig:coevo}.
The system incorporates an evolutionary search loop that employs idea tree-based solution generation to produce solutions in diverse formats. Additionally, it features a knowledge library that systematically stores and retrieves knowledge fragments throughout the search process.

\subsection{Idea Tree-based Solution Generation}
Solution generation is a crucial component of CoEvo, which is essential during both initialization and offspring production.
The framework employs an idea tree-based approach to generate solutions in multiple formats, following a three-step process: \textit{1)} inspiring, \textit{2)} thinking, and \textit{3)} solving.
Additionally, the solutions are generated in diverse formats to facilitate exploration across different search spaces.

\begin{figure}[t]
    \centering
    \includegraphics[width=0.38\textwidth]{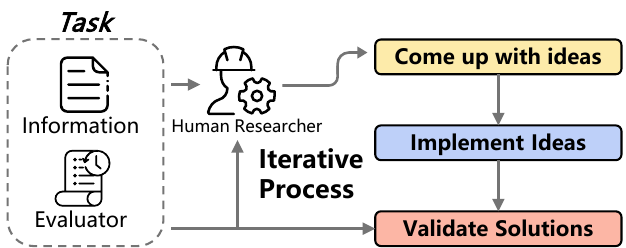}
    \caption{Human thinking process. It is usually an iterative process of idea generation, evaluation, and refinement.}
    \label{fig:human_think}
\end{figure}

Our three-step solution generation process closely mirrors human problem-solving, as illustrated in \figref{fig:human_think}.
Typically, humans first generate preliminary ideas when presented with a task and subsequently refine them.
This methodology aligns with two key insights from LLM research:
\textit{1)} the Reason-and-Act framework~\cite{yao:2023:synergizing}, which reflects the iterative cycle of reasoning and action, and \textit{2)} the tree-based multi-phase search process~\cite{wei:2022:chain,yao:2024:tree}, which promotes the exploration of diverse ideas.

Expanding on these insights, we propose an idea tree-based solution generation for task solution generation, as illustrated in \figref{fig:llm_think}.

\bheading{Idea Tree.}
The process begins by generating a diverse set of $N_0$ initial ideas, serving as the root nodes of the tree structure.
At each subsequent level $k$, $N_k$ ideas are developed through direct inference, guided by the evaluator’s feedback and existing ideas from the previous level.
This iterative refinement enhances the initial concepts and forms a network-like structure for deeper exploration.

Unlike the exhaustive sampling-and-branching process in Tree-of-Thought~\cite{yao:2024:tree}, our approach avoids exponential growth in computational resources by adopting a more constrained, network-like structure.
Nevertheless, the framework remains fully compatible with such extensions, ensuring scalability for future applications.

\bheading{Representations of Solutions.}
To facilitate the generation of solutions in diverse formats, we introduce multiple representations.
Prior research has demonstrated the effectiveness of parallel searches across natural language and Python code spaces, particularly in code generation tasks~\cite{wang:2024:planning,liu:2024:eoh,liu:2024:ael}.
Below, we present examples of several representative formats:
\begin{itemize}
    \item \textbf{Natural Language:} The foundational concept of LLMs is rooted in linguistic principles~\cite{zhao:2023:survey}. This domain aligns well with LLMs' capabilities and is central to related research.

    \item \textbf{Mathematical Formulas:} A fundamental representation for expressing mathematical functions and equations. In implementation, we use LaTeX code to represent these formulas.

    \item \textbf{Python Code:} This format is chosen because current LLMs are primarily trained on Python code, enhancing their code-generation capabilities~\cite{roziere:2023:code,dubey:2024:llama}.
          Additionally, code representations enable automated task evaluation~\cite{liu:2024:eoh,paredes:2024:funsearch}.

\end{itemize}

\begin{figure}[t]
    \centering
    \includegraphics[width=0.42\textwidth]{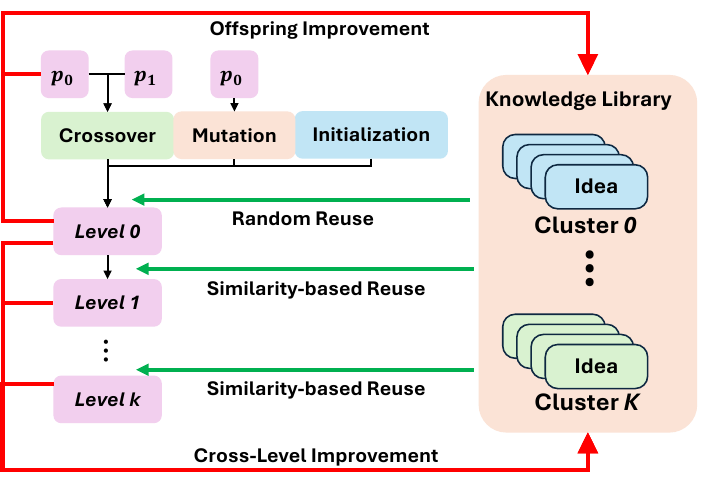}
    \vspace{-0.75\baselineskip}
    \caption{An illustration of the interaction between the knowledge library and the population. The red arrow represents the addition of knowledge to the library, while the green arrow denotes the reuse of knowledge.}
    \label{fig:knowledge_lib}
\end{figure}

\subsection{Knowledge Library}
\figref{fig:knowledge_lib} illustrates the interaction between the knowledge library and the population in generating ideas at different levels during the evolutionary search process.
Throughout this process, the library collects, stores, and reuses knowledge to enhance search efficiency.
To support these functions, the knowledge library incorporates three key mechanisms: \textit{1)} summarization, \textit{2)} management, and \textit{3)} reuse.

\bheading{Idea Summarization.}
Idea summarization occurs when solutions yield improved scores during tree-based search and offspring generation. An elevated score, as determined by the evaluator, signifies that the solution is effective and that the modifications introduced meaningful, knowledge-rich improvements.
Consequently, the LLM is prompted to extract and summarize the key idea underlying these changes. The summarized idea is then stored in the knowledge library in a structured definition-description format.

\bheading{Idea Management.}
Idea management is a critical component of the knowledge library, ensuring that stored ideas remain organized and retrievable.
We configure the knowledge library to maintain a finite number of ideas, as we want to avoid overwhelming the system with excessive knowledge and unleash the power of continual learning.
To prevent system overload and promote continual learning, the knowledge library maintains a finite number of ideas.
An excessive volume of ideas can hinder search and retrieval efficiency, and redundancy in learned knowledge cannot be guaranteed. To address this, we cluster ideas based on their semantic similarity, computed by cosine similarity between their sentence embeddings, thereby reducing redundancy while preserving key insights.

\bheading{Idea Reuse.}
The knowledge library is used to retrieve the knowledge when needed in two modes \emph{Random Reuse} and \emph{Similarity-based Reuse}.
When the LLM is generating new solutions, it needs to explore the search space to the largest extent, thus it is provided with random ideas from the knowledge library from each cluster for inspiration.
However, when it is conducting tree-based idea search, it needs to explore the ideas that are related to the ideas in the previous level, thus similarity-based reuse is used.
Similarity reuse involves calculate the distances from the current ideas to the ideas in the knowledge library and retrieve the ideas that are most similar to the current ideas.

\subsection{Evolutionary Search}
Our implementation of the evolution process adheres to the standard steps of evolutionary algorithms: initialization, crossover, mutation, and population management.

\bheading{Initialization.}
We initiate by randomly generating a set of $N$ solutions as the initial population using the tree-based idea search process in the previous section.
Notably, when the initialization starts with an empty knowledge library, the solutions are generated without any prior knowledge.
Otherwise the knowledge library is used to inspire the generation of initial solutions as described in \figref{fig:knowledge_lib}.

\bheading{Crossover.}
We employ two crossover operators—\emph{positive-crossover} and \emph{negative-crossover}.
Positive-crossover promotes the generation of solutions similar to parent ideas, whereas negative-crossover fosters the creation of distinctly different solutions, enhancing solution diversity as validated in related research~\cite{wang:2024:planning,liu:2024:autodan_turbo,liu:2024:eoh}.

\bheading{Mutation.}
We implement two mutation operators: \emph{positive-mutation} and \emph{negative-mutation}.
Positive-mutation introduces small, incremental changes to existing solutions, while negative-mutation applies more significant alterations.

\bheading{Population Update.}
We maintain the top $N$ solutions with the highest scores as the population for the next generation, ensuring a quality-driven evolution.

\section{Experiments}

\begin{table*}[t]
    \centering
    \caption{Comparison of the overall performance of our method with three categories of approaches: evolutionary search, deep learning, and LLM-SR. The normalized mean squared error (NMSE) is reported for both training (in-distribution, ID) and test (out-of-distribution, OOD) data. The best results are highlighted in bold with a gray background, while the second-best results are underlined.}
    \vspace{-0.75\baselineskip}
    \label{tab:overall_performance}
    \resizebox{0.98\textwidth}{!}{
        \begin{tabular}{llcccccccc}
            \toprule
            \multirow{2}{*}{\textbf{Category}}                & \multirow{2}{*}{\textbf{Method}}                          & \multicolumn{2}{c}{Oscillation 1}      & \multicolumn{2}{c}{Oscillation 2}      & \multicolumn{2}{c}{E. coli growth}      & \multicolumn{2}{c}{Stress-Strain}                                                                                                                                                                                 \\
            \cline{3-4} \cline{5-6} \cline{7-8} \cline{9-10}
                                                              &                                                           & ID $\downarrow$                        & OOD $\downarrow$                       & ID $\downarrow$                         & OOD $\downarrow$                        & ID $\downarrow$                               & OOD $\downarrow$                       & ID $\downarrow$                       & OOD $\downarrow$                       \\
            \midrule
            \multirow{2}{*}{\textbf{Evolutionary Search}$^*$} & GPlearn$^{[1]}$                                           & 0.0155                                 & 0.5567                                 & 0.7551                                  & 3.188                                   & 1.081                                         & 1.039                                  & 0.1063                                & 0.4091                                 \\
            \cline{2-10}
                                                              & PySR~\cite{cranmer:2023:interpretable}                    & 0.0009                                 & 0.3106                                 & 0.0002                                  & 0.0098                                  & 0.0376                                        & 1.0141                                 & 0.0331                                & 0.1304                                 \\
            \midrule
            \multirow{4}{*}{\textbf{Deep Learning}$^*$}       & NeSymReS~\cite{biggio:2021:nueral}                        & 0.0047                                 & 0.5377                                 & 0.2488                                  & 0.6472                                  & \multicolumn{2}{c}{\color{gray}{N/A ($d>3$)}} & 0.7928                                 & 0.6377                                                                         \\
            \cline{2-10}
                                                              & E2E~\cite{kamienny:2022:end}                              & 0.0082                                 & 0.3722                                 & 0.1401                                  & 0.1911                                  & 0.6321                                        & 1.4467                                 & 0.2262                                & 0.5867                                 \\
            \cline{2-10}
                                                              & DSR~\cite{petersen:2021:deep}                             & 0.0087                                 & 0.2454                                 & 0.0580                                  & 0.1945                                  & 0.9451                                        & 2.4291                                 & 0.3326                                & 1.108                                  \\
            \cline{2-10}
                                                              & uDSR~\cite{landajuela:2022:unified}                       & 0.0003                                 & 0.0007                                 & 0.0032                                  & 0.0015                                  & 0.3322                                        & 5.4584                                 & 0.0502                                & 0.1761                                 \\
            \midrule
            \multirow{5}{*}{\textbf{LLM-based}}               & LLM-SR (Mixtral)$^*$~\cite{shojaee:2024:llmsr}            & 7.89e-8                                & 0.0002                                 & 0.0030                                  & 0.0291                                  & 0.0026                                        & \underline{0.0037}                     & 0.0162                                & 0.0946                                 \\
            \cline{2-10}
                                                              & LLM-SR (\texttt{gpt-3.5-turbo})~\cite{shojaee:2024:llmsr} & 6.02e-9                                & 0.0004                                 & 2.55e-7                                 & 3.03e-3                                 & 0.0207                                        & 0.0547                                 & \underline{0.0017}                    & 0.0025                                 \\
            \cline{2-10}
                                                              & LLM-SR (\texttt{gpt-4o-mini})~\cite{shojaee:2024:llmsr}   & \underline{5.14e-9}                    & 0.0003                                 & \underline{1.79e-7}                     & \underline{3.11e-5}                     & 0.0214                                        & 0.0264                                 & 0.0020                                & 0.0020                                 \\

            \cline{2-10}
                                                              & \textbf{CoEvo (Ours, \texttt{gpt-3.5-turbo})}             & \cellcolor{light-gray}\textbf{4.32e-9} & \underline{8.71e-5}                    & \cellcolor{light-gray}\textbf{1.58e-10} & \cellcolor{light-gray}\textbf{1.32e-10} & \cellcolor{light-gray}\textbf{1.58e-9}        & \cellcolor{light-gray}\textbf{1.21e-8} & 0.0020                                & \underline{0.0015}                     \\
            \cline{2-10}
                                                              & \textbf{CoEvo (Ours, \texttt{gpt-4o-mini})}               & 1.28e-8                                & \cellcolor{light-gray}\textbf{7.51e-5} & 2.98e-7                                 & 2.21e-4                                 & \underline{0.0019}                            & 0.0107                                 & \cellcolor{light-gray}\textbf{0.0018} & \cellcolor{light-gray}\textbf{9.90e-4} \\

            \bottomrule
            \multicolumn{5}{l}{$^*$: Reported results from LLM-SR~\cite{shojaee:2024:llmsr}.}                                                                                                                                                                                                                                                                                                                                                                             \\
            \multicolumn{5}{l}{$^{[1]}$: \url{https://gplearn.readthedocs.io/en/stable/}}                                                                                                                                                                                                                                                                                                                                                                                 \\
            \\
        \end{tabular}
    }
    \vspace{-1.25\baselineskip}
\end{table*}

\subsection{Experimental Setup}
To evaluate the effectiveness of our proposed method, we conduct experiments on a subset of scientific problems from the AI Feynman benchmark~\cite{udrescu:2019:aifeynman} as described in LLM-SR\cite{shojaee:2024:llmsr}.
Specifically, we compare our method against leading symbolic regression techniques, including both evolutionary search and deep learning-based methods.
We also include LLM-SR, the most recent LLM-based symbolic regression method, as a baseline for LLM-driven approaches.
We present the experimental setup in the following subsections.

\bheading{Benchmark.}
We adopted the four problems introduced in LLM-SR~\cite{shojaee:2024:llmsr} for performance evaluation.
These problems are Oscillation 1, Oscillation 2, E. coli growth, and Stress-Strain.
They were specifically adapted from the original AI Feynman benchmark to prevent rote memorization of solutions. The AI Feynman benchmark~\cite{udrescu:2019:aifeynman}, which comprises 120 physics problems, is the current standard for evaluating symbolic regression methods in scientific equation discovery.

\bheading{Algorithms.}
We compare our method against a range of symbolic regression techniques, including both evolutionary search, deep learning-based methods, and LLM-based approaches.
For evolutionary search, we include \texttt{GPlearn}\footnote{\url{https://gplearn.readthedocs.io/en/stable/}} and \texttt{PySR}~\cite{cranmer:2023:interpretable}, which are two widely used symbolic regression libraries.
For deep learning-based methods, we consider NeSymReS~\cite{biggio:2021:nueral}, E2E~\cite{kamienny:2022:end}, DSR~\cite{petersen:2021:deep}, and uDSR~\cite{landajuela:2022:unified}.
For LLM-based approaches, we include LLM-SR~\cite{shojaee:2024:llmsr}, which is the most recent and advanced method in this category.
The evaluation results of non LLM-based methods are reported directly from LLM-SR~\cite{shojaee:2024:llmsr}.

For LLM-SR and our method, we limit the number of iterations to 2, 000 for fair comparison.
Moreover, we set the number of generations to 100 and the number of samples to 20 for each generation.
The size of the knowledge library is set to 30.


\bheading{Backbone LLMs.}
To assess the efficacy of LLM-based methods, we employ two backbone LLMs: \texttt{gpt-3.5-turbo} and \texttt{gpt-4o-mini}.
The former, \texttt{gpt-3.5-turbo}, is a widely adopted model known for its efficiency and effectiveness, serving as the default backbone in LLM-SR~\cite{shojaee:2024:llmsr}.
It offers a balance of cost-effectiveness and robust performance on symbolic regression tasks. To explore the potential of more advanced models, we also include \texttt{gpt-4o-mini}, a newer iteration with enhanced capabilities. This model is anticipated to excel in handling complex tasks.
We exclude other models, such as \texttt{Claude-3.5} (due to comparable performance to \texttt{gpt-3.5-turbo}) and \texttt{gpt-o1} (owing to prohibitive costs).

Specifically, \texttt{gpt-3.5-turbo} has a knowledge cutoff of September 2021~\cite{gpt35model}, while \texttt{gpt-4o-mini} extends to October 2023~\cite{gpt4model}.


\subsection{Overall Performance}
The overall performance of all methods on the selected benchmark problems is summarized in \tabref{tab:overall_performance}.
The normalized mean squared error (NMSE) serves as the performance metric, where lower values indicate better performance.

\bheading{Superior Performance.}
CoEvo consistently outperforms all other methods across all tested problems, achieving the lowest NMSE values.
This demonstrates its effectiveness in symbolic regression tasks.
Notably, CoEvo delivers results several orders of magnitude better on the Oscillation 2 and E. coli growth problems, highlighting its robustness in addressing complex symbolic regression challenges.

Our method exhibits minimal dependence on the choice of LLM, as it achieves comparable performance with both \texttt{gpt-3.5-turbo} and \texttt{gpt-4o-mini}.
Impressively, CoEvo successfully identifies the implicit equation for Oscillation 2, which is not discovered by LLM-SR using \texttt{gpt-4o-mini}.

\bheading{Characteristic of Oscillation 2.}
Oscillation 2 models the motion of an object governed by a precise physical equation defining its acceleration. Consequently, symbolic regression methods are expected to achieve near-zero NMSE by recovering the underlying analytical expression for acceleration.

However, the design of the evaluator presents a unique characteristic.
The evaluator processes data strictly in time order.
This sequential, time-series nature of the evaluation data introduces an alternative, highly effective pathway to compute acceleration: applying the \texttt{numpy.gradient} function directly to the velocity data yields an accurate numerical differentiation representing acceleration.

While other symbolic regression methods focused on recovering the explicit physical equation, only CoEvo, leveraging the capabilities of \texttt{gpt-3.5-turbo}, successfully discovered this implicit, data-driven formulation based on velocity differentiation.
This finding highlights CoEvo's unique ability to identify non-traditional, contextually optimal solutions that leverage the specific structure of the evaluation process.

\begin{figure}[t]
    \centering
    \subfigure[Oscillation 1]{\includegraphics[width=0.23\textwidth]{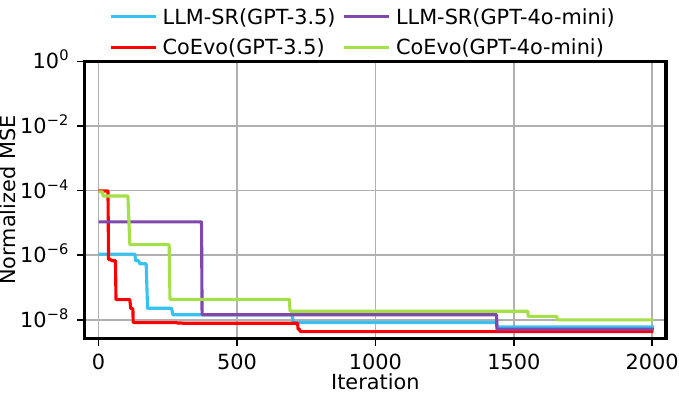}}
    \subfigure[Oscillation 2]{\includegraphics[width=0.23\textwidth]{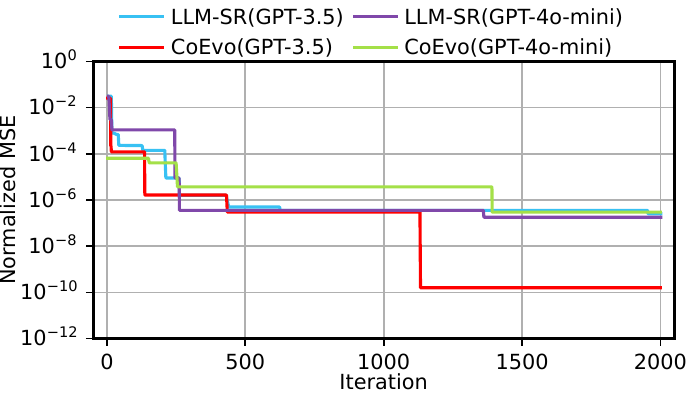}}
    \subfigure[E. coli growth]{\includegraphics[width=0.23\textwidth]{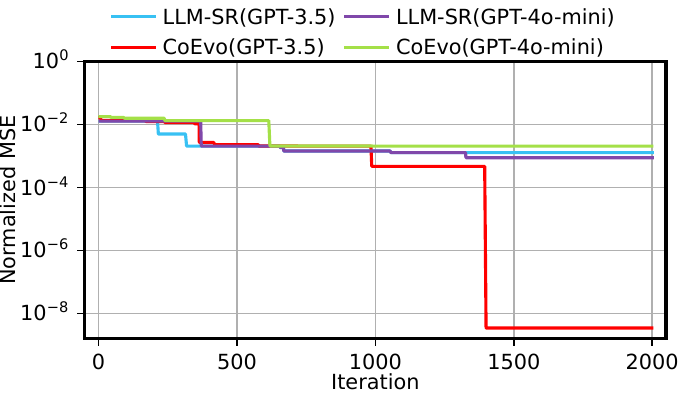}}
    \subfigure[Stress-Strain]{\includegraphics[width=0.23\textwidth]{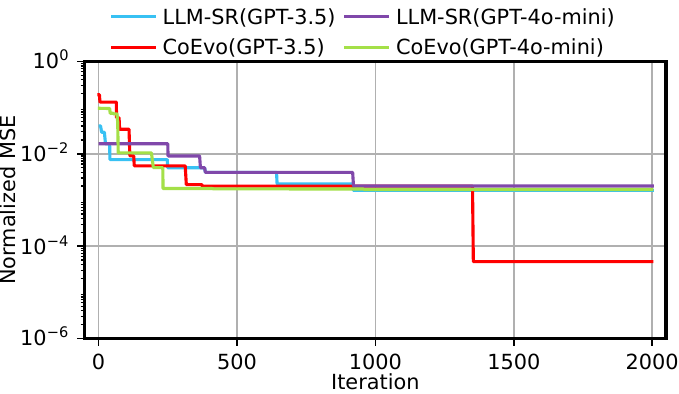}}
    \vspace{-0.75\baselineskip}
    \caption{The NMSE values during the search process of our method and LLM-SR.}
    \label{fig:search_process}
\end{figure}

\begin{figure}[t]
    \centering
    \subfigure[Oscillation 1]{\includegraphics[width=0.23\textwidth]{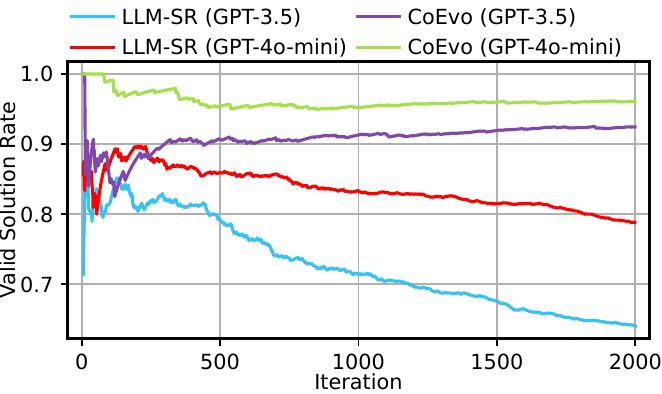}}
    \subfigure[Oscillation 2]{\includegraphics[width=0.23\textwidth]{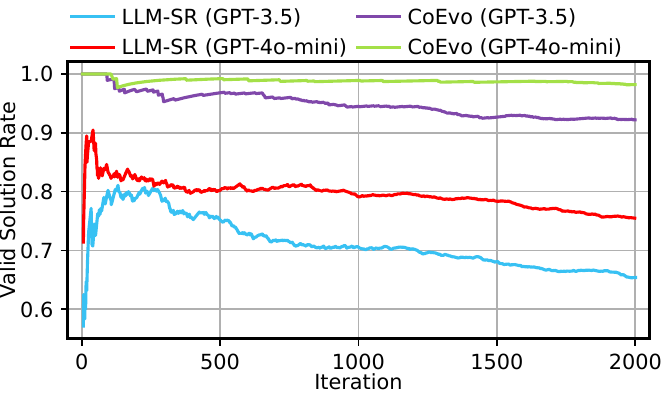}}
    \subfigure[E. coli growth]{\includegraphics[width=0.23\textwidth]{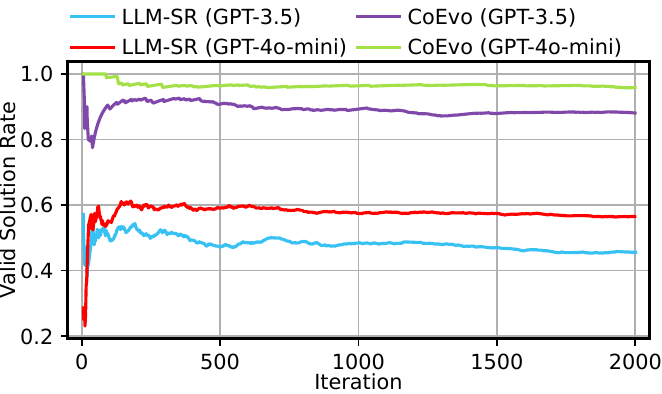}}
    \subfigure[Stress-Strain]{\includegraphics[width=0.23\textwidth]{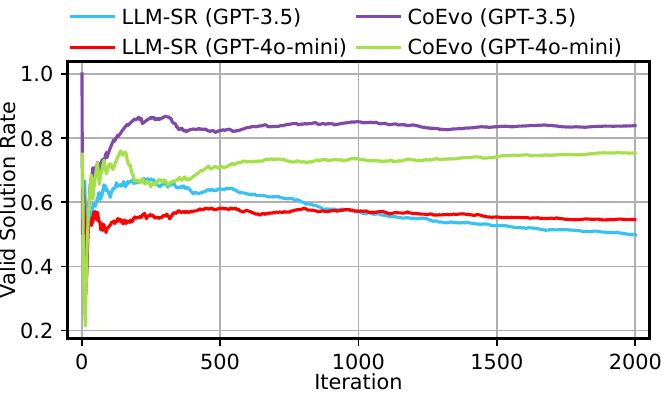}}
    \vspace{-0.75\baselineskip}
    \caption{The ratio of the generated solutions that are valid during the search process.}
    \label{fig:valid_ratio}
\end{figure}

\subsection{Convergence Analysis.}
In this section, we analyze the convergence behavior of our method compared to LLM-SR by examining the historical NMSE on training data and the ratio of valid solutions during the search process.
The historical NMSE reveals the convergence characteristics of the algorithms, while the ratio of valid solutions indicates the efficacy of the exploration strategies employed by each method.

\bheading{NMSE Convergence.}
\figref{fig:search_process} illustrates the historical NMSE across various benchmarks for our method and LLM-SR, using two distinct LLMs. The initial NMSE reduction (within the first 1,000 iterations) is similar across all cases, indicating that the ability of LLMs to generate valid solutions is comparable at the early stages of the search process.

When employing \texttt{gpt-3.5-turbo}, CoEvo achieves a significantly lower NMSE than both LLM-SR and its \texttt{gpt-4o-mini} counterpart. This result demonstrates that our method refines solutions more effectively over time and enables efficient exploration even with less advanced LLMs.

\bheading{Valid Solution Ratio.}
The ratio of valid solutions generated during the search process is illustrated in \figref{fig:valid_ratio}, demonstrating the exploration capabilities of our method and LLM-SR.
The valid solution ratio is defined as the proportion of error-free, evaluable solutions among all sampled solutions.

Overall, CoEvo produces significantly more valid solutions than LLM-SR across all benchmarks, indicating its superior effectiveness in exploring the search space. Notably, CoEvo achieves a higher valid ratio with \texttt{gpt-3.5-turbo} than LLM-SR does with \texttt{gpt-4o-mini}, demonstrating that our method can yield high-quality solutions even with less advanced LLMs.

An additional observation for both methods is that, for most benchmarks, the valid solution ratio increases with more powerful LLMs. The sole exception is the Stress-Strain benchmark for CoEvo.

\begin{figure}[t]
    \centering
    \subfigure[Oscillation 1]{\includegraphics[width=0.23\textwidth]{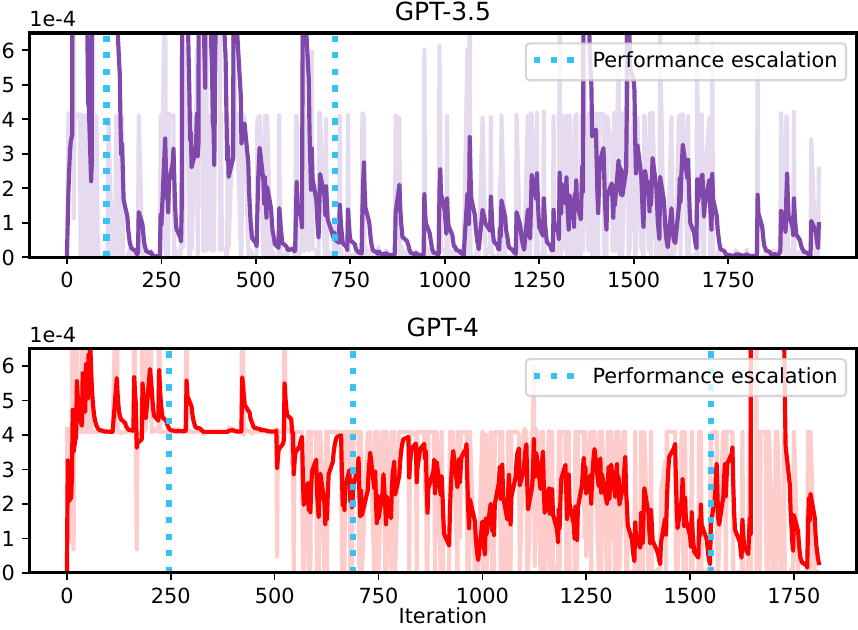}}
    \subfigure[Oscillation 2]{\includegraphics[width=0.23\textwidth]{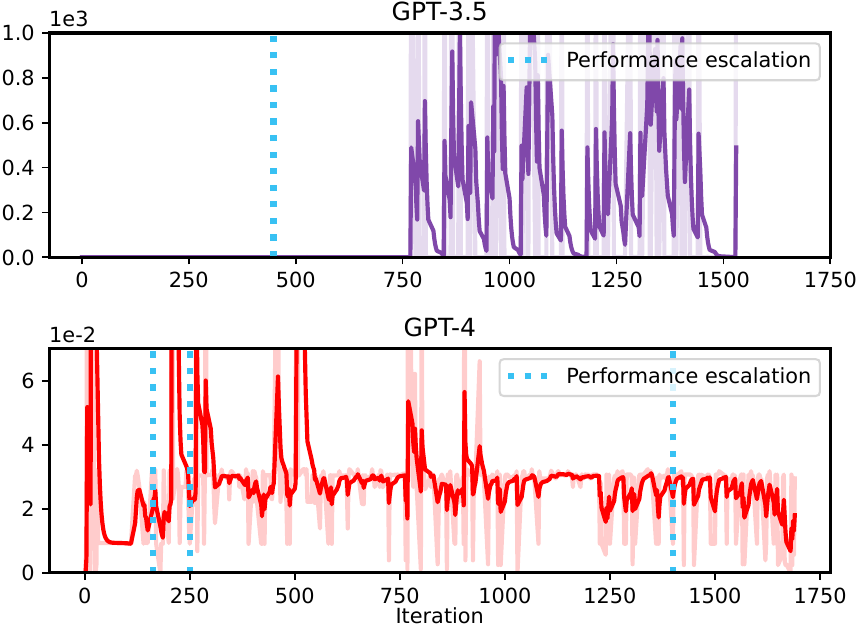}}
    \subfigure[E. coli growth]{\includegraphics[width=0.23\textwidth]{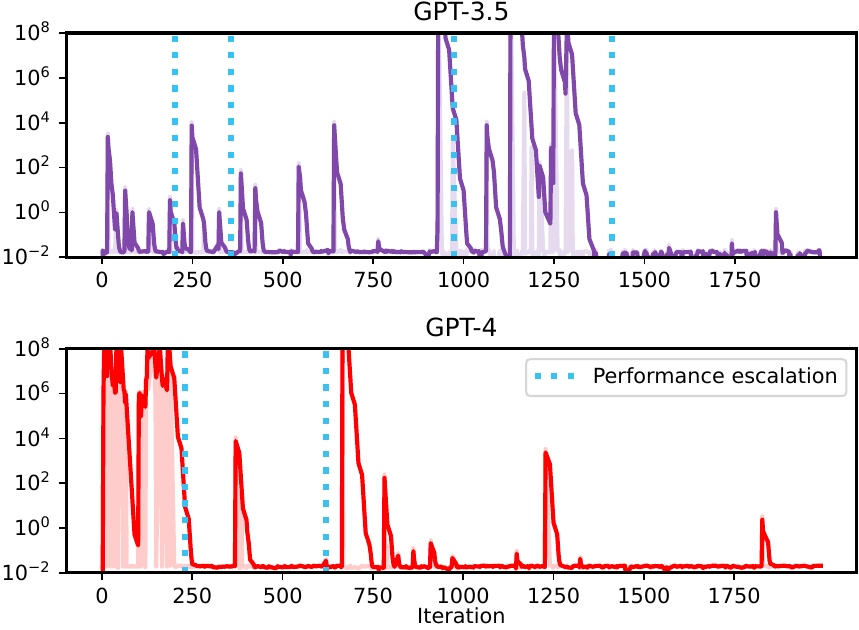}}
    \subfigure[Stress-Strain]{\includegraphics[width=0.23\textwidth]{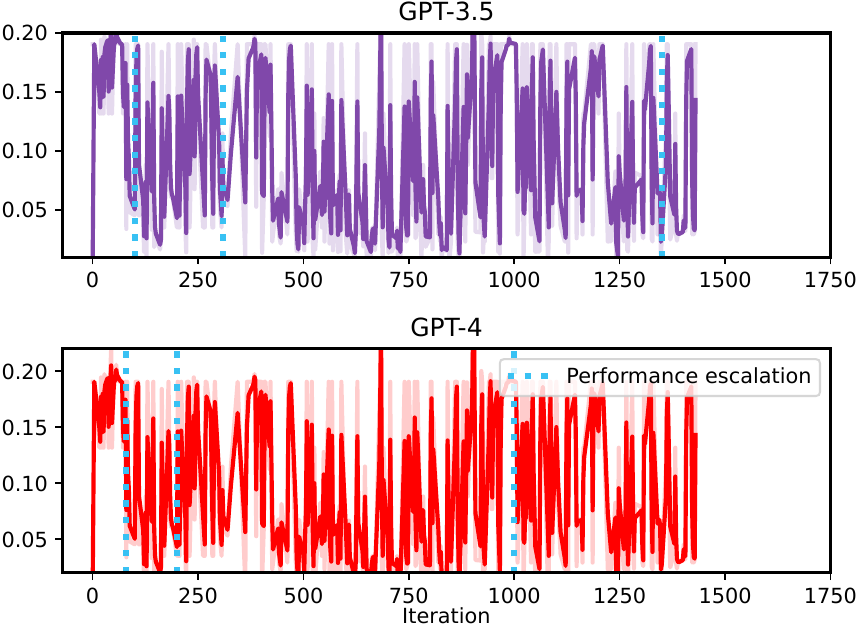}}
    \caption{The NMSE of the newly sampled solutions under the influence of the knowledge.}
    \label{fig:knowledge_influence}
\end{figure}

\begin{figure*}[t]
    \centering
    \subfigure[Oscillation 1+Knowledge]{\includegraphics[width=0.49\textwidth]{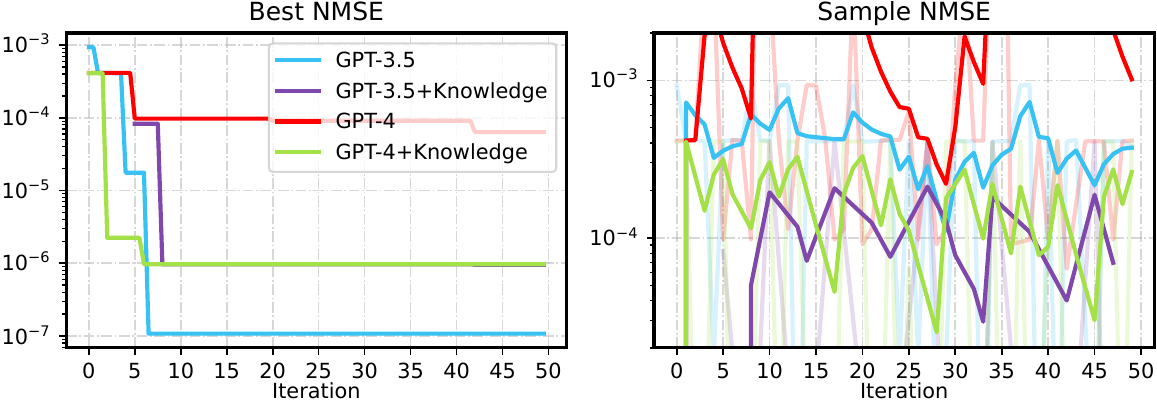}}
    \subfigure[E. coli growth+Knowledge]{\includegraphics[width=0.49\textwidth]{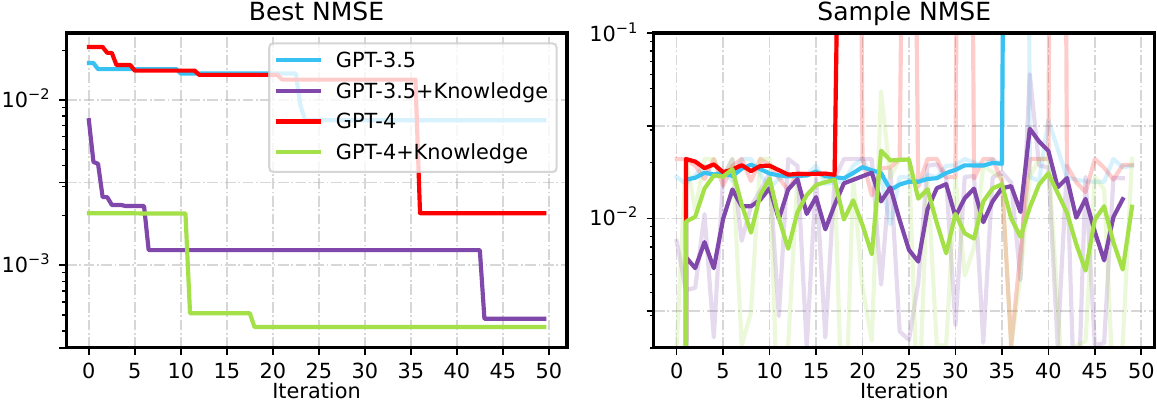}}
    \vspace{-0.75\baselineskip}
    \caption{Comparison of solution quality for Oscillation 1 and E. coli growth problems using knowledge extracted from \texttt{gpt-3.5-turbo} and \texttt{gpt-4o-mini}. Best NMSE convergence demonstrates the best NMSE achieved by the newly sampled solutions influenced by knowledge. Individual NMSE distributions show the performance of each solution.}
    \vspace{-1.5\baselineskip}
    \label{fig:use_knowledge}
\end{figure*}

\subsection{Knowledge Extraction and Application}
This section evaluates the impact of knowledge on the quality of solutions generated by CoEvo.
First, we examine solution quality across various benchmarks under the influence of dynamic knowledge managed by the knowledge library.
Next, we utilize knowledge extracted from different LLMs to generate new solutions and examine its effect on CoEvo's performance.
Finally, we visualize the knowledge extracted from Oscillation 2 to identify limitations of the knowledge.


\bheading{Solution Quality.}
\figref{fig:knowledge_influence} presents the NMSE of newly sampled solutions influenced by the knowledge library across different benchmarks over 2,000 iterations.
The performance escalation from \figref{fig:search_process} is indicated by blue dotted lines, representing the improvement of the best solution found by CoEvo.

A significant improvement in solution quality is observed in the Oscillation 1 and E. coli growth problems, where the NMSE of newly sampled solutions decreases by 2–3 orders of magnitude compared to the initial solutions. A similar improvement occurs in Oscillation 2 when using \texttt{gpt-4o-mini} knowledge. In contrast, the Stress-Strain problem is governed by a simple equation with empirically determined parameters, limiting the utility of knowledge; consequently, the NMSE of newly sampled solutions remains comparable to that of the initial solutions.

These improvements arise because the three problems are governed by fixed equations, allowing extracted knowledge to serve as informative hints for solution generation. However, in Oscillation 2, when the implicit function mentioned in the above subsection is identified, newly sampled solutions perform poorly. We further analyze this limitation in the next subsection by examining the underlying knowledge.

\bheading{Knowledge and Solution Generation.}
To isolate the influence of the knowledge library, we separately evaluate knowledge extracted from \texttt{gpt-3.5-turbo} and \texttt{gpt-4o-mini} to generate new solutions. Specifically, we collect knowledge extracted using both LLMs for the Oscillation 1 and E. coli growth problems and use them to generate 50 new solutions for each problem.

\figref{fig:use_knowledge} illustrates the best NMSE and the individual NMSE of the newly sampled solutions influenced by knowledge from \texttt{gpt-3.5-turbo} and \texttt{gpt-4o-mini}.
For individual NMSE, solutions generated with knowledge consistently outperform those without, exhibiting lower mean errors and reduced variability. For the best NMSE, knowledge-aided solutions converge more quickly, demonstrating that knowledge accelerates the discovery of better solutions.

\bheading{Investigation on Knowledge of Oscillation 2.}
Although CoEvo is the only method that identifies the implicit function for Oscillation 2, \figref{fig:knowledge_influence} shows that the NMSE of newly sampled solutions is significantly higher than that of the best solution found by CoEvo.
We examine the state of the knowledge library around the discovery of the implicit function and investigate the relationship between the knowledge and solution quality.

\begin{figure}[t]
    \centering
    \includegraphics[width=0.45\textwidth]{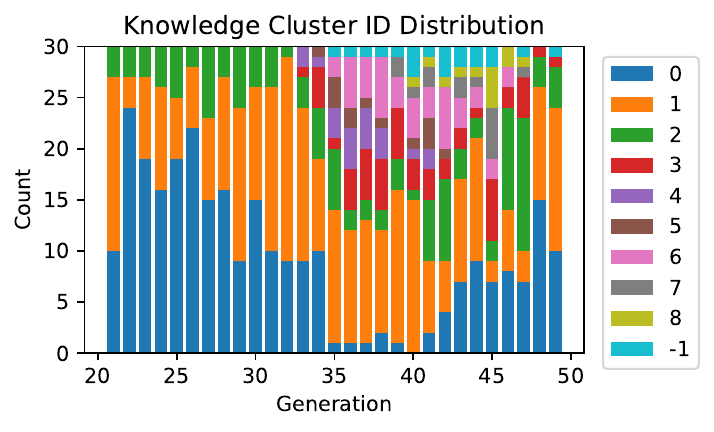}
    \vspace{-\baselineskip}
    \caption{Evolution of the knowledge library and idea distribution for Oscillation 2 (iterations 21-49).}
    \label{fig:knowledge_visual}
\end{figure}

\figref{fig:knowledge_visual} illustrates the knowledge library state for Oscillation 2 during the evolutionary search process from iteration 21 (400th sample) to iteration 50 (1000th sample). This interval is selected because it includes both high- and low-quality solutions guided by knowledge, as demonstrated in \figref{fig:use_knowledge}. Ideas from all chosen iterations are clustered using the DBSCAN algorithm, with distinct clusters marked by different colors.

The knowledge library evolves dynamically throughout the search process. Before the implicit function is discovered, the library contains fewer categories of useful knowledge, and solutions are concentrated in regions of the search space with low NMSE. After discovery, the library expands to encompass a broader range of useful knowledge, resulting in more diverse solutions with improved NMSE. This occurs because any knowledge incorporating the concept of \texttt{numpy.gradient} proves useful for solving the problem.

Based on this analysis, an idea condensation process may be necessary for the knowledge library to distill truly useful information.

\section{Conclusion}

Symbolic solutions are foundational to scientific and engineering innovation, yet their discovery remains constrained by the limitations of current methods in navigating vast representation spaces and evolving knowledge iteratively. To address this, we introduced CoEvo, a novel framework that reframes symbolic discovery as an open-ended, continual process. By synergizing LLMs with evolutionary search, CoEvo dynamically generates and refines symbolic solutions alongside a growing knowledge library.
CoEvo transcends the static reuse of prior knowledge typical of conventional LLM approaches, instead fostering genuine knowledge creation and refinement. This work pioneers a pathway toward human-like open-ended innovation in AI, where solutions and foundational understanding co-evolve endlessly to unlock novel scientific and engineering frontiers.

\section{Acknowledgments}
The work described in this paper was supported by the Research Grants Council of the Hong Kong Special Administrative Region, China [GRF Project No. CityU 11212524, 11217325].

\bibliography{main}

@book{boden:2004:creative,
	author    = {Boden, Margaret A},
	publisher = {Routledge},
	title     = {The creative mind: Myths and mechanisms},
	year      = {2004}
}

@book{brown:2000:fundamentals,
	author    = {Brown, Stephen D and Vranesic, Zvonko G},
	publisher = {McGraw-Hill New York},
	title     = {Fundamentals of digital logic with VHDL design},
	volume    = {70125910},
	year      = {2000}
}

@book{gielen:2012:symbolic,
	author    = {Gielen, Georges and Sansen, Willy MC},
	publisher = {Springer Science \& Business Media},
	title     = {Symbolic analysis for automated design of analog integrated circuits},
	volume    = {137},
	year      = {2012}
}

@book{wolf:2002:modern,
	author    = {Wolf, Wayne},
	publisher = {Pearson Education},
	title     = {Modern VLSI design: system-on-chip design},
	year      = {2002}
}

@article{abramson:2024:accurate,
	author    = {Abramson, Josh and Adler, Jonas and Dunger, Jack and Evans, Richard and Green, Tim and Pritzel, Alexander and Ronneberger, Olaf and Willmore, Lindsay and Ballard, Andrew J and Bambrick, Joshua and others},
	journal   = {Nature},
	publisher = {Nature Publishing Group UK London},
	title     = {Accurate structure prediction of biomolecular interactions with AlphaFold 3},
	year      = {2024}
}

@article{cranmer:2023:interpretable,
	author     = {Miles D. Cranmer},
	doi        = {10.48550/ARXIV.2305.01582},
	eprint     = {2305.01582},
	eprinttype = {arXiv},
	journal    = {CoRR},
	title      = {Interpretable Machine Learning for Science with PySR and SymbolicRegression.jl},
	volume     = {abs/2305.01582},
	year       = {2023}
}

@article{cranmer:2023:pysr,
	author     = {Miles D. Cranmer},
	eprinttype = {arXiv},
	journal    = {CoRR},
	title      = {Interpretable Machine Learning for Science with PySR and SymbolicRegression.jl},
	volume     = {abs/2305.01582},
	year       = {2023}
}

@article{dubey:2024:llama,
	author  = {Dubey, Abhimanyu and Jauhri, Abhinav and Pandey, Abhinav and Kadian, Abhishek and Al-Dahle, Ahmad and Letman, Aiesha and Mathur, Akhil and Schelten, Alan and Yang, Amy and Fan, Angela and others},
	journal = {arXiv preprint arXiv:2407.21783},
	title   = {The llama 3 herd of models},
	year    = {2024}
}

@article{huang:2024:surveyrag,
	author     = {Yizheng Huang and
	              Jimmy Huang},
	eprinttype = {arXiv},
	journal    = {CoRR},
	title      = {A Survey on Retrieval-Augmented Text Generation for Large Language
	              Models},
	year       = {2024}
}

@article{ifargan:2024:autonomous,
	author  = {Tal Ifargan  and Lukas Hafner  and Maor Kern  and Ori Alcalay  and Roy Kishony },
	doi     = {10.1056/AIoa2400555},
	eprint  = {https://ai.nejm.org/doi/pdf/10.1056/AIoa2400555},
	journal = {NEJM AI},
	number  = {0},
	pages   = {AIoa2400555},
	title   = {Autonomous LLM-Driven Research — from Data to Human-Verifiable Research Papers},
	volume  = {0},
	year    = {2024}
}

@article{lehman:2011:abandoning,
	author    = {Lehman, Joel and Stanley, Kenneth O},
	journal   = {Evolutionary computation},
	number    = {2},
	pages     = {189--223},
	publisher = {MIT Press},
	title     = {Abandoning objectives: Evolution through the search for novelty alone},
	volume    = {19},
	year      = {2011}
}

@article{lenat:1983:eurisko,
	author    = {Lenat, Douglas B},
	journal   = {Artificial intelligence},
	number    = {1-2},
	pages     = {61--98},
	publisher = {Elsevier},
	title     = {EURISKO: a program that learns new heuristics and domain concepts: the nature of heuristics III: program design and results},
	volume    = {21},
	year      = {1983}
}

@article{liu:2024:autodan_turbo,
	author  = {Xiaogeng Liu and
	           Peiran Li and
	           Edward Suh and
	           Yevgeniy Vorobeychik and
	           Zhuoqing Mao and
	           Somesh Jha and
	           Patrick McDaniel and
	           Huan Sun and
	           Bo Li and
	           Chaowei Xiao},
	journal = {CoRR},
	title   = {AutoDAN-Turbo: {A} Lifelong Agent for Strategy Self-Exploration to
	           Jailbreak LLMs},
	year    = {2024}
}

@article{liu:2024:systematic,
	author     = {Fei Liu and
	              Yiming Yao and
	              Ping Guo and
	              Zhiyuan Yang and
	              Zhe Zhao and
	              Xi Lin and
	              Xialiang Tong and
	              Mingxuan Yuan and
	              Zhichao Lu and
	              Zhenkun Wang and
	              Qingfu Zhang},
	doi        = {10.48550/ARXIV.2410.14716},
	eprint     = {2410.14716},
	eprinttype = {arXiv},
	journal    = {CoRR},
	title      = {A Systematic Survey on Large Language Models for Algorithm Design},
	volume     = {abs/2410.14716},
	year       = {2024}
}

@article{newell:1980:physical,
	author  = {Allen Newell},
	journal = {Cogn. Sci.},
	number  = {2},
	pages   = {135--183},
	title   = {Physical Symbol Systems},
	volume  = {4},
	year    = {1980}
}

@article{openai:2023:gpt4,
	author     = {OpenAI},
	eprint     = {2303.08774},
	eprinttype = {arXiv},
	journal    = {CoRR},
	title      = {{GPT-4} Technical Report},
	volume     = {abs/2303.08774},
	year       = {2023}
}

@article{paredes:2024:funsearch,
	author  = {Bernardino Romera{-}Paredes and
	           Mohammadamin Barekatain and
	           Alexander Novikov and
	           Matej Balog and
	           M. Pawan Kumar and
	           Emilien Dupont and
	           Francisco J. R. Ruiz and
	           Jordan S. Ellenberg and
	           Pengming Wang and
	           Omar Fawzi and
	           Pushmeet Kohli and
	           Alhussein Fawzi},
	journal = {Nat.},
	title   = {Mathematical discoveries from program search with large language models},
	year    = {2024}
}

@article{roziere:2023:code,
	author  = {Roziere, Baptiste and Gehring, Jonas and Gloeckle, Fabian and Sootla, Sten and Gat, Itai and Tan, Xiaoqing Ellen and Adi, Yossi and Liu, Jingyu and Sauvestre, Romain and Remez, Tal and others},
	journal = {arXiv preprint arXiv:2308.12950},
	title   = {Code llama: Open foundation models for code},
	year    = {2023}
}

@article{schmidt:2009:distilling,
	author    = {Schmidt, Michael and Lipson, Hod},
	journal   = {science},
	number    = {5923},
	pages     = {81--85},
	publisher = {American Association for the Advancement of Science},
	title     = {Distilling free-form natural laws from experimental data},
	volume    = {324},
	year      = {2009}
}

@article{shojaee:2024:llmsr,
	author     = {Parshin Shojaee and
	              Kazem Meidani and
	              Shashank Gupta and
	              Amir Barati Farimani and
	              Chandan K. Reddy},
	doi        = {10.48550/ARXIV.2404.18400},
	eprint     = {2404.18400},
	eprinttype = {arXiv},
	journal    = {CoRR},
	title      = {{LLM-SR:} Scientific Equation Discovery via Programming with Large
	              Language Models},
	volume     = {abs/2404.18400},
	year       = {2024}
}

@article{stanley:2017:open,
	author  = {Stanley, Kenneth O and Lehman, Joel and Soros, Lisa},
	journal = {While open-endedness could be a force for discovering intelligence, it could also be a component of AI itself},
	title   = {Open-endedness: The last grand challenge you’ve never heard of},
	year    = {2017}
}

@article{udrescu:2019:aifeynman,
	author     = {Silviu{-}Marian Udrescu and
	              Max Tegmark},
	eprint     = {1905.11481},
	eprinttype = {arXiv},
	journal    = {CoRR},
	title      = {{AI} Feynman: a Physics-Inspired Method for Symbolic Regression},
	volume     = {abs/1905.11481},
	year       = {2019}
}

@article{wang:2024:planning,
	author     = {Evan Wang and
	              Federico Cassano and
	              Catherine Wu and
	              Yunfeng Bai and
	              Will Song and
	              Vaskar Nath and
	              Ziwen Han and
	              Sean Hendryx and
	              Summer Yue and
	              Hugh Zhang},
	doi        = {10.48550/ARXIV.2409.03733},
	eprint     = {2409.03733},
	eprinttype = {arXiv},
	journal    = {CoRR},
	title      = {Planning In Natural Language Improves {LLM} Search For Code Generation},
	year       = {2024}
}

@article{wei:2022:chain,
	author  = {Wei, Jason and Wang, Xuezhi and Schuurmans, Dale and Bosma, Maarten and Xia, Fei and Chi, Ed and Le, Quoc V and Zhou, Denny and others},
	journal = {Advances in neural information processing systems},
	pages   = {24824--24837},
	title   = {Chain-of-thought prompting elicits reasoning in large language models},
	volume  = {35},
	year    = {2022}
}

@article{yao:2024:tree,
	author  = {Yao, Shunyu and Yu, Dian and Zhao, Jeffrey and Shafran, Izhak and Griffiths, Tom and Cao, Yuan and Narasimhan, Karthik},
	journal = {Advances in Neural Information Processing Systems},
	title   = {Tree of thoughts: Deliberate problem solving with large language models},
	volume  = {36},
	year    = {2024}
}

@inproceedings{biggio:2021:neural,
	author    = {Luca Biggio and
	             Tommaso Bendinelli and
	             Alexander Neitz and
	             Aur{\'{e}}lien Lucchi and
	             Giambattista Parascandolo},
	booktitle = {Proceedings of the 38th International Conference on Machine Learning,
	             {ICML}},
	pages     = {936--945},
	publisher = {{PMLR}},
	series    = {Proceedings of Machine Learning Research},
	title     = {Neural Symbolic Regression that scales},
	year      = {2021}
}

@inproceedings{biggio:2021:nueral,
	author    = {Luca Biggio and
	             Tommaso Bendinelli and
	             Alexander Neitz and
	             Aur{\'{e}}lien Lucchi and
	             Giambattista Parascandolo},
	booktitle = {Proceedings of the 38th International Conference on Machine Learning,
	             {ICML}},
	pages     = {936--945},
	publisher = {{PMLR}},
	series    = {Proceedings of Machine Learning Research},
	title     = {Neural Symbolic Regression that scales},
	volume    = {139},
	year      = {2021}
}

@inproceedings{kamienny:2022:end,
	author    = {Pierre{-}Alexandre Kamienny and
	             St{\'{e}}phane d'Ascoli and
	             Guillaume Lample and
	             Fran{\c{c}}ois Charton},
	booktitle = {Advances in Neural Information Processing Systems 35: Annual Conference
	             on Neural Information Processing Systems 2022, {NeurIPS}},
	title     = {End-to-end Symbolic Regression with Transformers},
	year      = {2022}
}

@inproceedings{kronberger2024inefficiency,
	author       = {Kronberger, Gabriel and Olivetti de Franca, Fabricio and Desmond, Harry and Bartlett, Deaglan J and Kammerer, Lukas},
	booktitle    = {International Conference on Parallel Problem Solving from Nature},
	organization = {Springer},
	pages        = {273--289},
	title        = {The inefficiency of genetic programming for symbolic regression},
	year         = {2024}
}

@inproceedings{landajuela:2022:unified,
	author    = {Mikel Landajuela and
	             Chak Shing Lee and
	             Jiachen Yang and
	             Ruben Glatt and
	             Cl{\'{a}}udio P. Santiago and
	             Ignacio Aravena and
	             Terrell Nathan Mundhenk and
	             Garrett Mulcahy and
	             Brenden K. Petersen},
	booktitle = {Advances in Neural Information Processing Systems 35: Annual Conference
	             on Neural Information Processing Systems 2022, {NeurIPS}},
	title     = {A Unified Framework for Deep Symbolic Regression},
	year      = {2022}
}

@inproceedings{liu:2024:ael,
	author    = {Fei Liu and
	             Xialiang Tong and
	             Mingxuan Yuan and
	             Xi Lin and
	             Fu Luo and
	             Zhenkun Wang and
	             Zhichao Lu and
	             Qingfu Zhang},
	booktitle = {Forty-first International Conference on Machine Learning, {ICML}},
	publisher = {OpenReview.net},
	title     = {Evolution of Heuristics: Towards Efficient Automatic Algorithm Design
	             Using Large Language Model},
	year      = {2024}
}

@inproceedings{liu:2024:eoh,
	author    = {Fei Liu and
	             Xialiang Tong and
	             Mingxuan Yuan and
	             Xi Lin and
	             Fu Luo and
	             Zhenkun Wang and
	             Zhichao Lu and
	             Qingfu Zhang},
	booktitle = {Forty-first International Conference on Machine Learning, ({ICML})},
	title     = {Evolution of Heuristics: Towards Efficient Automatic Algorithm Design
	             Using Large Language Model},
	year      = {2024}
}

@inproceedings{petersen:2021:deep,
	author    = {Brenden K. Petersen and
	             Mikel Landajuela and
	             T. Nathan Mundhenk and
	             Cl{\'{a}}udio Prata Santiago and
	             Sookyung Kim and
	             Joanne Taery Kim},
	booktitle = {9th International Conference on Learning Representations, {ICLR}},
	publisher = {OpenReview.net},
	title     = {Deep symbolic regression: Recovering mathematical expressions from
	             data via risk-seeking policy gradients},
	year      = {2021}
}

@inproceedings{yao:2023:synergizing,
	author    = {Shunyu Yao and
	             Jeffrey Zhao and
	             Dian Yu and
	             Nan Du and
	             Izhak Shafran and
	             Karthik R. Narasimhan and
	             Yuan Cao},
	booktitle = {The Eleventh International Conference on Learning Representations, {ICLR}},
	publisher = {OpenReview.net},
	title     = {ReAct: Synergizing Reasoning and Acting in Language Models},
	year      = {2023}
}

@misc{faldor:2024:omniepic,
	archiveprefix = {arXiv},
	author        = {Maxence Faldor and Jenny Zhang and Antoine Cully and Jeff Clune},
	eprint        = {2405.15568},
	primaryclass  = {cs.AI},
	title         = {OMNI-EPIC: Open-endedness via Models of human Notions of Interestingness with Environments Programmed in Code},
	url           = {https://arxiv.org/abs/2405.15568},
	year          = {2024}
}

@misc{gpt35model,
	author       = {OpenAI},
	howpublished = {\url{https://platform.openai.com/docs/models\#gpt-3-5-turbo}},
	note         = {Accessed: 2024-12-23},
	title        = {Models - OpenAI API}
}

@misc{gpt4model,
	author       = {OpenAI},
	howpublished = {\url{https://platform.openai.com/docs/models\#gpt-4o-mini}},
	note         = {Accessed: 2024-12-23},
	title        = {Models - OpenAI API}
}

@incollection{wigner:1990:unreasonable,
	author    = {Wigner, Eugene P},
	booktitle = {Mathematics and science},
	pages     = {291--306},
	publisher = {World Scientific},
	title     = {The unreasonable effectiveness of mathematics in the natural sciences},
	year      = {1990}
}

@misc{zhao:2023:survey,
	archiveprefix = {arXiv},
	author        = {Wayne Xin Zhao and Kun Zhou and Junyi Li and Tianyi Tang and Xiaolei Wang and Yupeng Hou and Yingqian Min and Beichen Zhang and Junjie Zhang and Zican Dong and Yifan Du and Chen Yang and Yushuo Chen and Zhipeng Chen and Jinhao Jiang and Ruiyang Ren and Yifan Li and Xinyu Tang and Zikang Liu and Peiyu Liu and Jian-Yun Nie and Ji-Rong Wen},
	eprint        = {2303.18223},
	primaryclass  = {cs.CL},
	title         = {A Survey of Large Language Models},
	url           = {https://arxiv.org/abs/2303.18223},
	year          = {2024}
}

\end{document}